\documentclass[sigconf]{acmart}

\AtBeginDocument{%
  }

\setcopyright{acmlicensed}
\copyrightyear{2018}
\acmYear{2018}
\acmDOI{XXXXXXX.XXXXXXX}

\acmConference[Conference acronym 'XX]{Make sure to enter the correct
  conference title from your rights confirmation emai}{June 03--05,
  2018}{Woodstock, NY}
\acmISBN{978-1-4503-XXXX-X/18/06}
\usepackage{amsmath,amsthm}

\usepackage{graphicx}
\usepackage{algorithmic}
\usepackage{algorithm}
\usepackage{textcomp}
\usepackage{soul}
\usepackage{array}
\usepackage{amsmath}
\usepackage{cases}
\usepackage{tipa}
\usepackage{mathrsfs}
\usepackage{graphicx}
\usepackage{subfig}
\usepackage{bbding}
\usepackage{pifont}
\usepackage{multirow}
\usepackage{amsmath}

\begin{document}

\title{Binary Linear Tree Commitment-based Ownership Protection for Distributed Machine Learning}

\author{Tianxiu Xie}
\email{3120215672@bit.edu.cn}
\affiliation{%
  \institution{Beijing Institute of Technology}
  \country{China}
}

\author{Keke Gai}
\email{gaikeke@bit.edu.cn}
\affiliation{%
  \institution{Beijing Institute of Technology}
  \country{China}
}

\author{Jing Yu}
\email{yujing02@iie.ac.cn}
\affiliation{%
  \institution{Institute of Information Engineering, CAS}
  \country{China}
}

\author{Liehuang Zhu}
\email{liehuangz@bit.edu.cn}
\affiliation{%
  \institution{Beijing Institute of Technology}
  \country{China}
}


\begin{abstract}
Distributed machine learning enables parallel training of extensive datasets by delegating computing tasks across multiple workers. 
Despite the cost reduction benefits of distributed machine learning, 
the dissemination of final model weights often leads to potential conflicts over model ownership as workers struggle to substantiate their involvement in the training computation.
To address the above ownership issues and prevent accidental failures and malicious attacks, verifying the computational integrity and effectiveness of workers becomes particularly crucial in distributed machine learning.
In this paper, we proposed a novel binary linear tree commitment-based ownership protection model to ensure computational integrity with limited overhead and concise proof.
Due to the frequent updates of parameters during training, 
our commitment scheme introduces a maintainable tree structure to reduce the costs of updating proofs. 
Distinguished from SNARK-based verifiable computation, our model achieves efficient proof aggregation by leveraging inner product arguments.
Furthermore, proofs of model weights are watermarked by worker identity keys to prevent commitments from being forged or duplicated. 
The performance analysis and comparison with SNARK-based hash commitments validate the efficacy of our model in preserving computational integrity within distributed machine learning.
\end{abstract}

\ccsdesc[500]{Security and privacy~Digital rights management}
\ccsdesc[500]{Computing methodologies~Neural networks}
\ccsdesc[300]{Computing methodologies~Distributed algorithms}
\keywords{Distributed Identity Audit, Deep Neural Network, Model Ownership Verification, Blockchain, Intellectual Property Protection}

\maketitle

\section{Introduction}

High-performance machine learning models typically involve expensive iterative optimization, owing mainly to the requirements for voluminous labeled data and extensive computing resources. To reduce overall training costs, \textit{Distributed Machine Learning} (DML) allows model owners to outsource training tasks to multiple workers \cite{zhou2023dynamic}.
Until the model converges, each worker is required to submit their local model weights to the model owner every epoch. 
In the current context, one of the primary issues in DML may be model ownership disputes. First, some model architectures are publicly available, so once the final parameters of DML are released, it is a challenge for workers in the training process to prove their contribution to the final model. 
Moreover, beyond the imperative of preserving the privacy of model weights, the model owner requires to verify the computational integrity and validity performed by workers.
In the absence of integrity protection, malicious entities may manipulate intermediate variables during the model training phase or expropriate model weights within an untrusted cloud environment (in Figure \ref{fig1}).
Consequently, to defend against accidental worker failure (e.g., hardware crashes) and Byzantine attacks (e.g., data and model poisoning attacks) \cite{cao2022mpaf, shan2022poison}, it matters for workers to prove the computational integrity and reliability to confirm model ownership. 

\begin{figure}[!t]
    \centering
        \centering
	\includegraphics[width=0.9\linewidth]{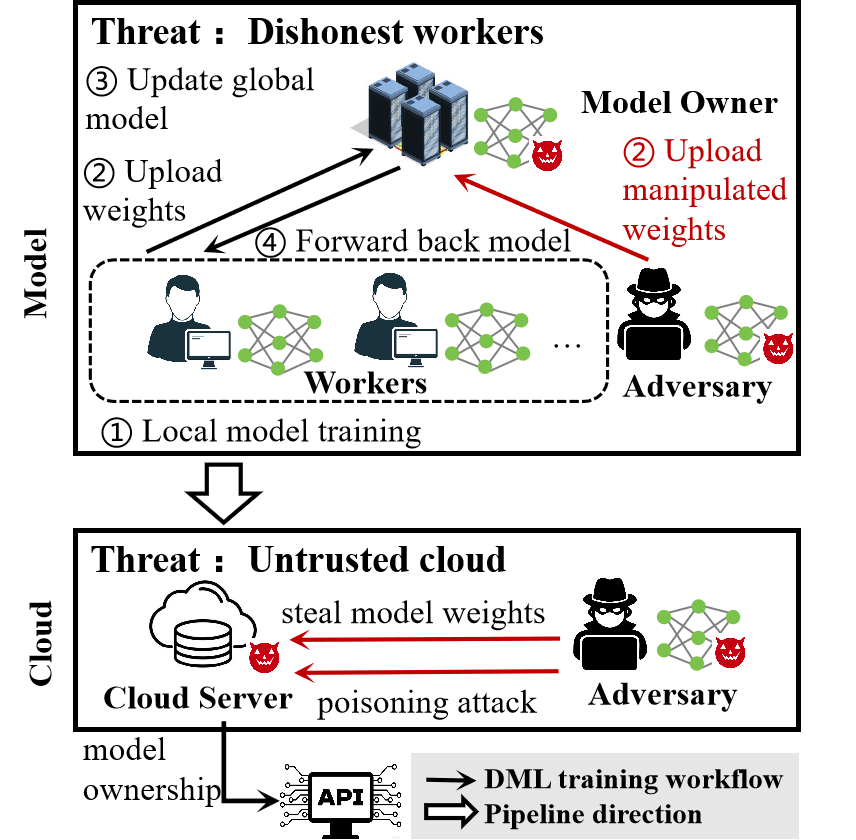}
        \label{fig1}
\caption{The potential threat on distributed machine learning. For distributed model training, an adversary may upload manipulated weight.
For an untrusted cloud, an adversary may steal model weights from honest workers or conduct a poisoning attack
}
\label{fig1}
\end{figure}

%

A typical scheme of proving computational integrity is using cryptographic primitives \cite{ben2014succinct}. 
For example, verifiable computation or security commitment (e.g., SNARK and STARK) requires workers to provide additional evidence alongside the training execution results \cite{albrecht2022lattice, dong2023rai2}. 
The model owner can verify the provided proofs to ensure the training tasks needed for the final model are executed properly.
In addition to non-interactive arguments, interactive proof systems  guarantee the integrity of the reasoning process \cite{miller2020adversarial}.
However, 
the overhead associated with opening and verifying these proofs can be substantial, often introducing an overwhelming overhead. 
Model parameter iteration involves frequent updating of these proofs, so that this method is costly due to the proof maintenance. 
Furthermore, 
\textit{Proof-of-Learning} (PoL) scheme is proposed to prove computational integrity \cite{jia2021proof}. 
\textit{Proof-of-Learning} (PoL) involves workers' intermediate models during training, along with the corresponding data points from the training set used for model updating. 
However, \cite{zhang2022adversarial} generated a PoL spoof with reduced computational and storage costs by using ``adversarial examples". 
Therefore, a critical challenge in DML is to find out an effective proof scheme with limited overhead and concise proof during the training process. 
Effective proofs need to satisfy two primary properties as follows:
\begin{enumerate}
    \item The workers have expended effort and computational resources to execute the training tasks;
    \item Training tasks have been computed correctly, indicating the workers have maintained computational integrity.
\end{enumerate}



To address the challenges above, we propose a \textit{Binary Linear Tree Commitment-Based DML Ownership Protection} (BLTC-DOP) model. 
Distributed learning iteratively updates model weights (i.e., intermediate variables for model training) over a series of vectors through a stochastic gradient-based optimization (e.g., SGD) \cite{woodworth2020minibatch}. 
The vector of model weights represents the necessary optimization computations for workers to train the model, denoted by $\mathbf{W}$ $=$ $(w_{0},w_{1},w_{2},...)$. 
A worker generates a commitment $C$ for vector $\mathbf{W}$ and generates a commitment certificate. 
Commitment $C$ can guarantee the integrity and validity of weight information without disclosing specific model weights. 
The model weights are iterated by workers who perform calculations at each training round.
When the weight vector changes, the commitment scheme of BLTC-DOP leverages the binary linear tree structure to update all proofs in commitment $C$ in sub-linear time instead of computing the commitment from scratch. 
To prevent the adversary from forging or copying commitment $C$, the identity key of the worker is used as the watermark to associate the commitment $C$ with 
the worker. 
The trusted model owner verifies the validity of the commitment watermark and certificates. The certificates generated by workers are stored on a distributed ledger(e.g., blockchain). The timestamp of the distributed ledger can ensure the both spatial and temporal properties of the certificate to protect the copyright of the final model.


Our main contributions are summarized as follows:
\begin{enumerate}
    \item We propose a novel commitment-based scheme for model ownership protection. Our proposed model retains the original internal structure and the training optimization algorithms, only introducing additional commitments to guarantee computational integrity. We limit the time cost of commitment verification and update through a multi-linear tree structure, thereby realizing an efficient commitment scheme that is both aggregatable and maintainable.
     \item BLTC-DOP ensures the non-stealability of commitments without compromising the operational effectiveness of the DML process. As the commitment embeds the worker's identity watermark, it becomes immune to forgery by adversaries. The watermarked proof does not affect the verification efficiency, aggregation efficiency and update efficiency of the commitment.
     \item The proposed model outperforms state-of-the-art SNARK-based approaches on distributed training, which demonstrates the feasibility and effectiveness of the proposed model.
\end{enumerate}
\section{Methodology} \label{sec:mod}


\subsection{Overview of BLTC-DOP} 
Our proposed BLTC-DOP model can verify the computational integrity of the final DML model training process. 
We remark that the DML training process usually involves overwhelming computational overhead, so BLTC-DOP considers the training process rather than the inference process. 
Compared with changing the training structure and iterative optimization algorithm of DML, BLTC-DOP only requires workers to make additional commitments to the model weights. 
To be specific, our approach can address two following issues.
(1) The model owner verifies the correctness of the training result $f^{a}_{W_{P}}$ submitted by the worker in each round ($f^{a}_{W_{P}}$ represents the model trained locally by the worker);
(2) One of the workers claims its ownership of the final DML model $f_{W_{P}}$.


We assume that the worker participating in distributed learning is prover $\mathcal{P}$, the model owner is verifier $\mathcal{V}$, and the dishonest or malicious worker is adversary $\mathcal{A}$. The worker generates a commitment $C$ to the model weight vector $\mathbf{W}$.
During the DML training process, $\mathcal{P}$ reveals to $\mathcal{V}$ a commitment certificate \verb|Cert|$(\mathcal{P},f_{W_{P}},C)$ with its significant, where $f_{W_{P}}$ represents the model trained by $\mathcal{P}$, and $C$ represents the commitment value of the model weight vector generated by $\mathcal{P}$. 
During the DML training process, $\mathcal{P}$ discloses the model intermediate variables at intervals, and synchronously updates the commitment value in the corresponding certificate. To defend against replay attacks (i.e., $\mathcal{A}$ claims to have submitted the certificate first), $\mathcal{V}$ checks the certificate stored on the distributed ledger and verifies the validity of the signature.
Specifically, from model initialization to final model training, $\mathcal{V}$ can confirm the validity of the intermediate variables trained by $\mathcal{P}$ through commitment $C$ at any time. This in turn proves that $\mathcal{P}$ performed the training work required to obtain the final model. 
In addition, as mentioned in the background, because model training involves a large number of weight parameter updates, Commitment $C$ also requires expensive maintenance and verification costs to ensure the computational integrity of workers.
Computational integrity and validity means that during the calculation process of distributed training, the correctness of the calculation can be ensured through proof verification.
In the process of distributed training, computing integrity means that the worker has completely and correctly performed the distributed training tasks, thereby ensuring the worker's unique training process.
The BLTC-DOP model introduces a multi-linear tree structure to simplify the update and aggregation process of commitment $C$ so that $\mathcal{P}$ can effectively verify $C$ in sub-linear time. 
Moreover, $\mathcal{A}$ may use its knowledge about the model structure, parameters, and datasets to forge certificates to deceive $\mathcal{V}$.
Our proposed commitment scheme also uses identity watermarks to distinguish certificates forged by adversaries.


\subsection{Security in BLTC-DOP}\label{threat}
In order to seize ownership of the model, adversary $\mathcal{A}$ aims to deceive verifier $\mathcal{V}$ by falsely claiming that it performed the computation required to train $f_{W_{P}}$. Because $\mathcal{A}$ does not spend the necessary computing resources and does not train the model $f_{W_{P}}^{a}$, $\mathcal{A}$ hopes to construct a fake certificate \verb|Cert|$(\mathcal{A},f_{W_{P}},C)$ to forge its own training work. We assume that adversary $\mathcal{A}$ has unlimited access to the model structure, model weights, and user public keys, and consider $\mathcal{A}$'s ability to effectively deceive $\mathcal{V}$ from the following three aspects:
\begin{enumerate}
    \item[\textbf{M1.}] Adversary $\mathcal{A}$ aims to remove the cryptographic signature on the \verb|Cert|$(\mathcal{P},f_{W_{P}},C)$ and replace it with its own signature, thereby stealing $\mathcal{P}$’s certificate or creating a \verb|Cert| $(\mathcal{A},$  $f_{W_{P}},$  $C)$ that is exactly the same as  \verb|Cert|$(\mathcal{P},f_{W_{P}},C)$, i.e., \verb|Cert|$(\mathcal{A},f_{W_{P}},C)$ $=$ \verb|Cert|$(\mathcal{P},f_{W_{P}},C)$.
    \item[\textbf{M2.}] Adversary $\mathcal{A}$ aims to forge an invalid certificate for $f_{W_{P}}$, but it can be verified by $\mathcal{V}$. $\mathcal{A}$ generates two inconsistent certificates for the same model weight vector. The certificate created by $\mathcal{A}$ may be different from $\mathcal{P}$'s certificate, i.e., \verb|Cert|$(\mathcal{A},f_{W_{P}},C)$ $\neq$ \verb|Cert|$(\mathcal{P},f_{W_{P}},C')$. Conflicting certificates may cause honest $\mathcal{P}$'s certificate verification to fail.
    \item[\textbf{M3.}] Adversary $\mathcal{A}$ may attack the centralized cloud to manipulate the global model weights, and cause wrong training involving workers at specified data points, so that the correct credentials cannot be verified.
\end{enumerate}


\begin{algorithm}[!t]
	\caption{Binary Linear Tree-based Commitment Scheme}
	\label{alg:1}
	\begin{algorithmic}[1]
		\REQUIRE{the commitment $C$ and the proof $\pi$ of the $\mathbf{W}$}	
            \ENSURE{security parameter $\lambda$, the model weight vector $\mathbf{W}$ $=$ $(w_{0},w_{1},...,w_{l-1})$ 
            , the length $l$ of $\mathbf{W}$}
        \STATE{\emph{/* Phase 1 : Proof Generation for BLTC */}}
        \STATE{$pp$ $\gets$ $\mathsf{BLTC.Gen}(\lambda, l)$: Inputs the security parameter $\lambda$ and the length $l$ and outputs the public parameters $pp$. The public parameters $pp$ can be called by all other algorithms.}
        \STATE{($WMK$, $PVK$) $\gets$ $\mathsf{BLTC.Mark}(\lambda)$: Outputs a randomly generated key pair, which includes a private watermarking key $WMK$ and a public verification key $PVK$.}
        \STATE{$wmpp$ $\gets$ $\mathsf{BLTC.MarkPPGen}(pp, WMK)$: Inputs the public parameters $pp$ and the private watermarking key $WMK$, and outputs a watermarked public parameters $wmpp$.}
        \STATE{$C$ $\gets$ $\mathsf{BLTC.Com}_{pp}(\mathbf{W})$: Inputs the vector $\mathbf{W}$ $=$ $(w_{0},w_{1},...,w_{l-1})$ and outputs the commitment $C$ of $\mathbf{W}$ $\in$ $\mathbb{Z}^{l}_{p}$.}
        \STATE{$\pi_{i}$ $\gets$ $\mathsf{BLTC.Open}_{wmpp}(w_{i},i)$: Inputs the the position $i$ and the message $w_{i}$ at the $i$-th position in $\mathbf{W}$. The algorithm outputs the corresponding proof $\pi_{i}$, which is computed by the watermarked public parameters $wmpp$ under $WMK$.}
        \STATE{$(\pi_{0},\pi_{1},...,\pi_{l-1})$ $\gets$ $\mathsf{BLTC.OpenAll}_{wmpp}(\mathbf{W})$: Outputs all watermarked proof $\pi_{i}$ for $\mathbf{W}$.}
        \STATE{\emph{/* Phase 2 : Proof Aggregation for BLTC */}}
        \STATE{$\pi_{U}$ $\gets$ $\mathsf{BLTC.Aggr}_{pp}(w_{i},\pi_{i},U)$ ($i \in U$): Inputs the individual vector value $w_{i}$ and the corresponding proof $\pi_{i}$ and aggregates them into one proof $\pi_{U}$. The aggregated proof $\pi_{U}$ can be used to verify the \textit{Correctness} and \textit{Soundness} of $\mathbf{W}$.}
        \STATE{\emph{/* Phase 3 : Proof Verification for BLTC */}}
        \STATE{$(0/1)$$\mathsf{BLTC.Verify}_{pp}(C,U,\pi_{U},w_{i}, PVK_{i})$ ($i \in U$): Inputs the proof $\pi_{U}$, the commitment $C$ and the vector value $w_{i}$. The algorithm outputs $0$ (i.e., $w_{i}$ is not the $i$-th message of $\mathbf{W}$) or $1$ (i.e., $w_{i}$ is the $i$-th message of $\mathbf{W}$). In addition, $\mathcal{V}$ verifies the watermarks of the proof $\pi_{U}$ via $PVK_{i}$.}
        \STATE{\emph{/* Phase 4 : Proof Update for BLTC */}}
        \STATE{$C'$ $\gets$ $\mathsf{BLTC.ComUpdate}_{pp}(C,i',\xi)$ ($\xi \in \mathbb{Z}_{p}$): Outputs the updated commitment $C'$ computed by the the $i'$-th position with $\xi$. $\xi$ represents the change value of the vector $w_{i'}$ at the $i'$-th position.}
        \STATE{$(\pi_{0}',\pi_{1}',...,\pi_{l-1}')$ $\gets$ $\mathsf{BLTC.ProofAllUpdate}_{wmpp}(i',\xi,(\pi_{0},$ $\pi_{1},...,\pi_{l-1}))$ ($\xi \in \mathbb{Z}_{p}$): Outputs the updated watermarked proof $(\pi_{0}',\pi_{1}',...,\pi_{l-1}')$ computed by the the $i'$-th position with change value $\xi$.}
	\end{algorithmic}
\end{algorithm}

Refer to the threat model and the two properties of effective proof, we promise that the scheme should satisfy the following properties:
\begin{enumerate}
    \item[\textbf{P1.}] \textit{Correctness}: The prover $\mathcal{P}$ obtains this \verb|Cert| by training the model from random initialization of model parameters until they converge to $f_{W_{P}}$, then the \verb|Cert| of $f_{W_{P}}$ should be verifiable successfully. 
    We assume that both $\mathcal{P}$ and $\mathcal{V}$ are honest. Specifically, during the training process from model initialization to $f_{W_{P}}$, $\mathcal{P}$ follows the commitment process to perform correct calculations, then the commitment $C$ of $\mathcal{P}$ can verify the correctness with a significant degree of certainty.
    \item[\textbf{P2.}] \textit{Soundness}: We assume that $\mathcal{A}$ has dishonest or malicious behavior, then the probability of the commitment $C$ generated by $\mathcal{A}$ passing the verification of $\mathcal{V}$ is negligible. That is, it is difficult for $\mathcal{A}$ to impersonate an honest $\mathcal{P}$, and $\mathcal{A}$ cannot cheat with a conflicting certificate.
    \item[\textbf{P3.}] \textit{Aggregability}: $\mathcal{P}$ can effectively aggregate multiple proofs $\pi_{i} (i \in U)$ in a commitment into a brief proof $\pi_{U}$. This property can reduce the verification cost of $\mathcal{V}$ and improve the verification efficiency through proof aggregation.
    \item[\textbf{P4.}] \textit{Maintainability}: When the model weight vector changes, rather than recomputing the commitment with linear time, maintainability requires that all proofs in the commitment be updated in sub-linear time. This property guarantees bounded generation and update promise overhead.
\end{enumerate}


\subsection{Binary Linear Tree-based Commitment Scheme}\label{commitment}
To ensure the \textit{Aggregability} and \textit{Maintainability}, we introduce the binary linear tree structure in VC. 
A trapdoor commitment scheme possesses both binding and hiding properties. Binding ensures that the commitment is uniquely associated with a particular original value, thereby preventing deception or alteration of the corresponding original value. Hiding guarantees that the commitment does not disclose any information regarding the original value, making it infeasible to derive the original value from the commitment value.
Our proposed Binary Linear Tree-based Commitment (BLTC) scheme involves four main phases, which are \textbf{Proof Generation}, \textbf{Proof Aggregation}, \textbf{Proof Verification} and \textbf{Proof Update}.  
In BLTC, $\mathcal{P}$ computes the commitment $C$ and proof $\pi$ of the model weight vector $\mathbf{W}$, while $\mathcal{V}$ verifies the proof according to the commitment.
Algorithm \ref{alg:1} shows the structure of our BLTC scheme.

\textbf{Phase 1: Proof Generation for BLTC.} 
Without loss of generality, we assume that the length of $\mathbf{W}$ is $l=$ $2^{n}$ and $\mathbf{W}$ $=$ $(w_{0},w_{1},...,w_{l-1})$ $\in$ $\mathbb{Z}^{l}_{p}$. Refer to the \textit{Hyperproof} \cite{srinivasan2022hyperproofs} and \textit{Polynomial Commitments} \cite{zhang2022multi}, the vector value $w_{i}$ at the $i$-th position of $\mathbf{W}$ can be represented by a multilinear extended polynomial function $f:$ $\mathbb{Z}^{n}_{p}$ $\to$ $\mathbb{Z}_{p}$ as shown in Eq. (\ref{eq:1}):
\begin{equation}\label{eq:1}
    f(\mathbf{i}) = f(i_{n},i_{n-1},...,i_{1}) = w_{i}
\end{equation}
In order to extend $f$ from $w_{i}$ to $\mathbf{W}$, we assume a trapdoor $\mathbf{A}$ $=$ $(a_{n},a_{n-1},...,a_{1})$ $\in_{R}$ $\mathbb{Z}^{n}_{p}$ and define the selection functions $S_{k_{j}}$ and $Y_{k,n}$ ($k$ $\in$ [$0$,$2^{n})$) as follows:
\begin{equation}\label{eq:2}
    S_{k_{j}}(a_{j})= 
        \begin{cases}
           a_{j},   & k_{j}=1 \\
           1-a_{j},  & k_{j}=0  
        \end{cases}
\end{equation}
\begin{equation}\label{eq:3}
    Y_{k,n}(\mathbf{A}) = \prod^{n}_{j=1}S_{k_{j}}(a_{j})
\end{equation}
We note that $Y_{0,0}(\mathbf{A})=1$. According to Eq. (\ref{eq:1}), Eq. (\ref{eq:2}) and Eq. (\ref{eq:3}), the multilinear extended polynomial function $f$ of $\mathbf{W}$ is as follows:
\begin{equation}\label{eq:4}
    \begin{split}
        f(\mathbf{A}) = \sum^{l-1}_{k=0}w_{k}Y_{k,n}(a_{n},a_{n-1},...,a_{1})
        =\sum^{l-1}_{k=0}w_{k}Y_{k,n}(\mathbf{A})
    \end{split}
\end{equation}

The BLTC generates the commitment $C$ via a bilinear pairing group, which denotes generating groups $\mathbb{G}_{1}$ and $\mathbb{G}_{2}$ with generators $g_{1}$ and $g_{2}$, respectively, and a pairing $e:$ $\mathbb{G}_{1}$ $\times$ $\mathbb{G}_{2}$ $\to$ $\mathbb{G}_{T}$. We assume that $g^{t_{j}-i_{j}}_{2}$($j \in [0,n]$) is the verification key of the $i$-th position in $\mathbf{W}$. Eq.\ref{eq:5} as follows is the commitment $C$ computed via $g_{1}$, $\mathbf{W}$ and $\mathbf{A}$ (line 5 in Alg.\ref{alg:1}):
\begin{equation}\label{eq:5}
    \begin{split}
        C &= g_{1}^{f(\mathbf{A})} = g_{1}^{f(a_{n},a_{n-1},...,a_{1})}\\
          &= g_{1}^{\sum^{l-1}_{k=0}w_{k}Y_{k,n}(\mathbf{A})}
          =\prod^{n-1}_{j=0}(g_{1}^{Y_{k,n}(\mathbf{A})})^{w_{k}}
    \end{split}
\end{equation}

\begin{figure}
    \centering
    \includegraphics[width=1.0\columnwidth]{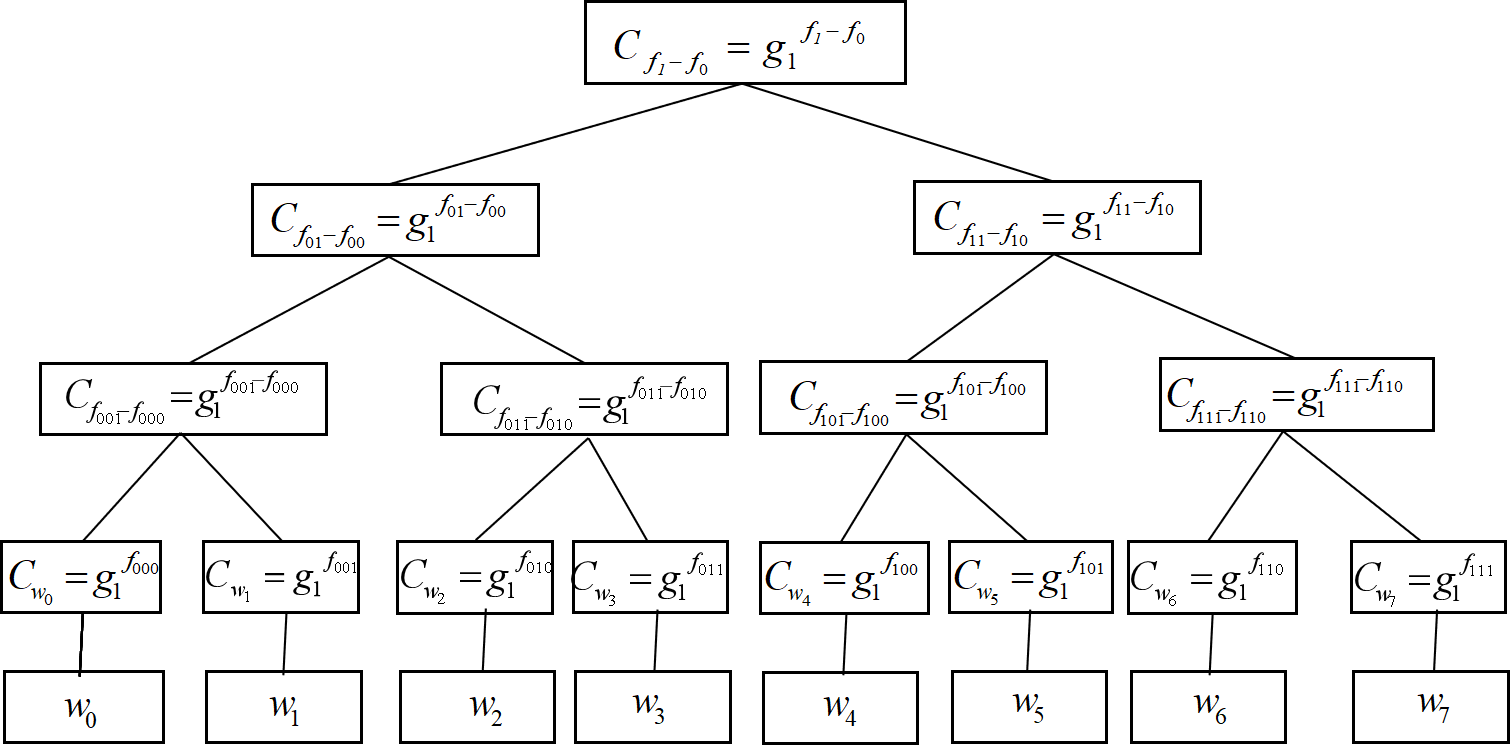}
    \caption{An example of the binary linear tree with the size of 8. Each node represents a commitment. All commitments involved in the path to root of $w_{i}$ represents the proof for $w_{i}$.}
    \label{fig2}
\end{figure}

We reduce the $\mathsf{BLTC.Open(\cdot)}$ time to $O(nlogn)$ via a binary linear tree. The binary linear tree is generated by the decomposition of the multilinear extended polynomial function $f$. First, We divide $f$ by $a_{n}-t$ ($t \in \left\{0,1\right\}$) into two multilinear extended polynomial functions $f_{0}$ and $f_{1}$ such that $f=(f_{1}-f_{0}) \cdot (a_{n}-t) + f_{t}$. Next, we generate a commitment $g_{1}^{(f_{1}-f_{0})(\mathbf{A})}$. Finally, we recursively disperse the polynomial function (i.e., disperse $f_{0}$ and $f_{1}$ into two polynomial functions as well) and compute the commitment until $f$ is a constant. Figure \ref{fig2} shows the structure of the binary linear tree in BLTC. Specifically, $(\pi_{0},\pi_{1},...,\pi_{l-1})$ can be represented by the binary linear tree (line 7 in Alg.\ref{alg:1}), while $\pi_{i}$ is represented by the $i$-th path in the binary linear tree (line 6 in Alg.\ref{alg:1}), denoted for $\pi_{i}=(x_{i,n},x_{i,n-1},...,x_{i,1})$. We note that $x_{i,n}$ is the $n$-th node of the $i$-th path in the binary linear tree. 

Binary linear tree is a divide-and-conquer algorithm that decomposes a polynomial into smaller-scale sub-problems and leverages the relationships between these sub-problems to perform recursive computations.  
This process ultimately transforms the polynomial from its coefficient representation to a point-valued representation.
By analyzing the recursive calculation, we can express its time complexity as $T(n)$ = $2T(n/2)$. Applying the Master Theorem to this equation, we can conclude that the overall time complexity of the $\mathsf{BLTC.Open(\cdot)}$ algorithm is $O(nlogn)$.

\textbf{Phase 2: Proof Aggregation for BLTC.} 
Inspired by \textit{Hyperproofs}, our BLTC scheme uses \textit{Non-interactive Inner Product Arguments} to aggregate proofs (\textbf{P3}). To aggregate a series of individual proofs $\pi_{i}$ into a proof $\pi_{U}$, we assume that vector $\mathbf{R}=(\mathbf{R}_{1}||\mathbf{R}_{2}||...||\mathbf{R}_{m}, m =|U|)$ denotes proofs $\pi_{i}(i \in U)$ and $\mathbf{T}=($ $\mathbf{T}_{1}||$ $\mathbf{T}_{2}||$ $...$ $||\mathbf{T}_{m}, $ $m =|U|)$ denotes the corresponding verification keys, where $\mathbf{R},\mathbf{T} \in \mathbb{G}_{1}^{m} \times \mathbb{G}_{2}^{m}$. For the sake of brevity, we define the $\mathsf{BLTC.Aggr}$ algorithm as follows:\\
$\mathsf{BLTC.Aggr_{pp}}((w_{i},\pi_{i},U))$:
\begin{enumerate}
    \item the pairing product $Z= \left\langle \mathbf{R},\mathbf{T} \right\rangle=\prod_{b=1}^{m}e(\mathbf{R}_{b},\mathbf{T}_{b})$.
    \item a commitment key pairing $ComKey=(sk,pk) \in \mathbb{G}_{1}^{m} \times \mathbb{G}_{2}^{m}$ used to computed $B \gets \left\langle \mathbf{R}, sk \right\rangle$.
    \item $v_{b} \gets H(B,\mathbf{T},b) \in \mathbb{Z}_{p}$ where $H(\cdot)$ is a random oracle function and $b \in [1,m]$.
    \item compute $\mathbf{T'} \gets (\mathbf{T}_{1}^{v_{1}}||\mathbf{T}_{2}^{v_{2}}||...||\mathbf{T}_{m}^{v_{m}})$ for $m =|U|$.
    \item return $\pi_{U} \gets (B, \mathsf{(IPA.Compute}(ComKey,\mathbf{R},\mathbf{T'})))$, where the $\mathsf{IPA.Compute}$ algorithm is used to compute the proof $\pi$ that Aggregated Commitment $AC$ $\gets$ $\mathsf{IPA.Com}(ComKey;$ $\mathbf{R},$ $\mathbf{T},$ $\langle \mathbf{R},\mathbf{T} \rangle)$  
\end{enumerate}

\textbf{Phase 3: Proof Verification for BLTC.} 
We assume that $\theta$ is the Private Watermarking Key ($WMK$) of $\mathcal{P}$ and we embed it into the proof $\pi$ as follows (line 4 in Alg.\ref{alg:1}):
\begin{equation}
    \pi_{i}^{\theta}=(x_{i,n}^{\theta},x_{i,n-1}^{\theta},...,x_{i,1}^{\theta})
\end{equation}
The corresponding Public Verification Key ($PVK$) is $g_{2}^{\theta}$ (line 3 in Alg.\ref{alg:1}). $\mathcal{V}$ verifies the effectiveness of proof $\pi_{i}^{\theta}$ by computing $e(C/g_{1}^{a_{i}},g_{2}^{\theta})$ and $\prod_{j \in [0,n]}e(x_{i,j}^{\theta},g_{2}^{a_{j}-i_{j}})$, respectively (line 11 in Alg.\ref{alg:1}). The $\mathsf{BLTC.Verify}$ algorithm would output $1$ in the case where Eq. (\ref{eq:7}) holds as follows (i.e., the proof $\pi_{U}$ is effective):
\begin{equation}\label{eq:7}
    e(C/g_{1}^{a_{i}},g_{2}^{\theta})=\prod_{j \in [0,n]}e(x_{i,j}^{\theta},g_{2}^{a_{j}-i_{j}})
\end{equation}

\begin{table*}[]
    \centering
    \begin{tabular}{ccccccc}
        \hline
        \multicolumn{2}{c}{Theoretical Performance} & Merkle  & Verkle  & Merkel SNARK & Pointproofs & Our scheme  \\
        \hline
        \multicolumn{1}{c}{\multirow{3}{*}{Proof Size}} & Individual proof size & $O($log$l)$ & $O($log$_{b}l)$ & $O($log$l)$  & $O(1)$ & $O($log$l)$\\
        \multicolumn{1}{c}{} & Aggregate proof size & $\times$ & $\times$ & $O(1)$ & $O(1)$  & $O($log$(u$log$l))$\\
        \multicolumn{1}{c}{} & Commitment size & $O(1)$ & $O(1)$ & $O(1)$  &$O(1)$ & $O(1)$\\
        \hline
        \multicolumn{1}{c}{\multirow{3}{*}{Time costs}} & Generate proofs time & $O(l)$ & $O(bl)$ & $O(l)$ & $O(l$log$l)$ &  $O(l$log$l)$\\
        \multicolumn{1}{c}{}&  Aggregate proofs time & $\times$ & $\times$ & $O(u$log$l$log$(u$log$l))$  &  $O(u)$ & $O(u$log$l)$\\
        \multicolumn{1}{c}{}& Verify time (for agg. proof) & $\times$ & $\times$ & $O($log$l)$ & $O(u)$ & $O(u$log$l)$ \\
        \multicolumn{1}{c}{} &  Update time (for agg. proof) & $\times$ & $\times$ & $O($log$l)$ & $O(l)$ &  $O($log$l)$ \\
        \hline
        \multicolumn{1}{c}{Ownership} & \multirow{2}{*}{Non-stealability} & \multirow{2}{*}{$\times$} & \multirow{2}{*}{$\times$} & \multirow{2}{*}{$\times$} & \multirow{2}{*}{$\checkmark$} & \multirow{2}{*}{$\checkmark$}  \\
        \multicolumn{1}{c}{Protect} &  &  &  &  & &  \\
        \hline
    \end{tabular}
    \caption{Comparison with other verifiable computing schemes, which can not simultaneously support \textit{Aggregability} (``Aggregate proofs time" in line 6), \textit{Maintainability} (``Update time" in line 8) and \textit{Non-stealability} (line 9). The size of the weight vector is $l=2^{n}$, where $n$ is the bit-width of weight vector elements (we noted that $O(n)$ group operations are counted as an exponentiation). The number of aggregated individual proofs is $u$. Proof sizes are in terms of group elements, while time complexities are in term of group exponentiations or field operations. Both Verify time and Update time are the time costs for aggerated proofs.}
    \label{tab:1}
\end{table*}

\textbf{Phase 4: Proof Update for BLTC.}
We assume that the vector value $w_{i'}$ (at position $i'$ in $\mathbf{W}$) changes by $\xi$ after a training epoch (\textbf{P4}). Some auxiliary information $aux$ for position $i'$ is required to update the Commitment $C$. Refer to Eq.s (\ref{eq:2}) and (\ref{eq:3}), the $aux_{i'}$ for position $i'$ is as follows:
\begin{equation}
        aux_{i'} =\left\{ aux_{i',j}, j \in [0,n] \right\}
        =\left\{ g_{1}^{Y_{i',j}(\mathbf{A})}, j \in [0,n] \right\}
\end{equation}

Thus for any $j \in [0,n]$, $\pi_{i'}=(x_{i',n},x_{i',n-1},...,x_{i',1})$ of the $i'$-th path can be updated as follows  (line 14 in Alg.\ref{alg:1}):
\begin{equation}
    x_{i',j}' = x_{i',j} \cdot (aux_{i',j-1})^{\xi}
    =x_{i',j} \cdot (g_{1}^{Y_{i',j-1}(\mathbf{A})})^{\xi}
\end{equation}
In addition, as for the proof $\pi_{i}$ of $w_{i}$ at position $i\neq i'$, we consider the same nodes where the $i$-th path and the $i'$-th path intersect on the binary linear tree. Specifically, we assume that $i$ and $i'$ have $b$ same bits, which means for any $j \in [n-b+1,n]$, $i_{j} = i'_{j}$. Thus the proof $\pi_{i}$ of $w_{i}$ can be updated as follows:
\begin{equation}
    x_{i,j}' = x_{i,j} \cdot (aux_{i,j-1})^{\xi}
    =x_{i,j} \cdot (g_{1}^{Y_{i,j-1}(\mathbf{A})})^{\xi}
\end{equation}
The commitment $C$ is updated as (line 13 in Alg. \ref{alg:1}):
\begin{equation}
    C' = C \cdot g_{1}^{Y_{i',n}(\mathbf{A})}
    = C \cdot (aux_{i',n})^{\xi}
\end{equation}




\section{Theoretical Analysis} \label{sec:sa}


\subsection{Security Analysis}
\textbf{Correctness of BLTC.} \textit{Correctness} denotes the property that the commitment computed by any honest worker is verifiable with a significant degree of certainty. When using a vector commitment scheme, correctness is guaranteed when opening a commitment value. Specifically, when a user wants to expose one or some of the committed values in a vector commitment, the commitment verifier can verify the consistency of these open values with their original committed values without relying on the proofs provided by the prover. Specifically, all algorithms in BLTC should be executed correctly, which means that model weights watermarked with honestly-generated private watermarking keys should be successfully verified (\textbf{P1}). \textit{Correctness} formally requires as follows:

\begin{subequations}\label{eq:8}
\small
    \begin{align}
        &Pr\left[ \mathsf{BLTC.Verify}_{pp}(C,i,\pi_{i},w_{i}, PVK_{i}) = 1 \right]=1.\\ 
        &Pr\left[ \mathsf{BLTC.Verify}_{pp}(C,U,\pi_{U},w_{i}, PVK_{i}) = 1 \right]=1.\\
        &Pr\left[ \mathsf{BLTC.Verify}_{pp}(C',U,\pi_{U}',w_{i'}', PVK_{i}) = 1 \right]=1
    \end{align}
\end{subequations}

Eq. (\ref{eq:8}a), Eq. (\ref{eq:8}b) and Eq. (\ref{eq:8}c) denote the \textit{Correctness} of individual proof $\pi_{i}$, aggregated proof $\pi_{U}$ and updated proof $\pi_{U}'$, respectively. We remark that proofs generated via $\mathsf{BLTC.Open(\cdot)}$ or $\mathsf{BLTC.OpenAll(\cdot)}$ are successfully verified by $\mathsf{BLTC.Verify(\cdot)}$ in the \textbf{Proof Generation}, \textbf{Proof Aggregation}, and \textbf{Proof Update} of BLTC. The certificates are stored on the distributed ledger, so attacker cannot fail the verification of certificates from honest workers by attacking the central cloud (\textbf{M3}).

\textbf{Soundness of BLTC.} \textit{Soundness} ensures that the model owner $\mathcal{V}$ cannot be cheated in the case of an adversary attempting to forge conflicting certificates \verb|Cert|$(\mathcal{A},f_{W_{p}},C)$ or invalid certificates (\textbf{M2}). Namely, it is required that for each \textit{Probabilistic Polynomial Time} (PPT) algorithm $\mathscr{A}$, the chance of winning BLTC is negligible:
\begin{equation}\label{eq:6}
    \resizebox{0.48\textwidth}{!}{$
    Pr\left[
    \begin{matrix}
        pp \gets \mathsf{BLTC.Gen}(\lambda, l),\\
        \left( C,
        \begin{matrix}
        (U, w_{i}, PVK_{i},\pi_{U})_{i \in U},\\
        (V, w_{j}', PVK_{j},\pi_{V}')_{j \in V}
        \end{matrix}
        \right) \gets \mathscr{A}(\lambda,pp):\\
        1 \gets \mathsf{BLTC.Verify}_{pp} \left(
        \begin{matrix}
            C,U,\pi_{U},\\
            w_{i}, PVK_{i} 
        \end{matrix}
        \right)_{i \in U} \wedge\\
        1 \gets \mathsf{BLTC.Verify}_{pp} \left(
        \begin{matrix}
            C,V,\pi_{V}',\\
            w_{j}', PVK_{j} 
        \end{matrix}
        \right)_{j \in V} \wedge\\
        \exists t \in U \cap V, w_{t} \neq w_{t}'
    \end{matrix}
    \right] \leq \mathrm{negl(\lambda)}.
    $}
\end{equation}
Our proposed BLTC scheme is sound (\textbf{P2}) in the case where Eq. (\ref{eq:6}) holds, as BLTC can prevent $\mathcal{A}$ from generating inconsistent certificates for $w_{t} \neq w_{t}'$ at the same position $t$.

\textbf{Non-stealability of BLTC.} \textit{Non-stealability} means that an adversary $\mathcal{A}$ who knows the $PVK$ but not the $WMK$ would not be able to claim that this $\mathcal{A}$ performs the training tasks. 
In order to prevent adversaries from removing cryptographic signatures, our scheme incorporates a watermark directly into the commitment, ensuring the non-stealability of the data.
In BLTC, the attack for $\mathcal{A}$ to steal the commitment is to remove $\theta$ from the watermarked proof $\pi_{i}^{\theta}$. However, this attack requires the exponent value $\theta^{-1}$, which $\mathcal{A}$ does not know. 
BLTC scheme correlates the commitment with the user's identity by embedding $\theta$ of $WMK$ into the proof. 
Therefore, $\mathcal{A}$ cannot steal the certificate by removing or replacing the $\mathcal{P}$'s signature (\textbf{M1}).


\begin{table*}[]
\centering
\begin{tabular}{ccccccccccc}
\hline
\multirow{2}{*}{Time Costs}         & \multicolumn{10}{c}{$n$ =30}                                                                 \\ \cline{2-11} 
& $u$=4 & $u$=8 & $u$=16 & $u$=32 & $u$=64 & $u$=128 & $u$=256 & $u$=512 & $u$=1024 & $u$=2048 \\ \hline
Aggregate proofs time (s)           & 0.04  & 0.08  & 0.16   & 0.32   & 0.64   & 1.26    & 2.48    & 4.93    & 9.83     &   19.85       \\ \hline
Verify an individual proof time (s) & 0.02  & 0.03  & 0.06   & 0.12   & 0.24   & 0.47    & 0.94    & 1.89    & 3.77     &   7.66       \\ \hline
Verify an aggregated proof time (s) & 0.04  & 0.07  & 0.13   & 0.25   & 0.49   & 0.95    & 1.82    & 3.15    & 6.64     &    12.76      \\ \hline
Update commitment time (s)          & 0.02  & 0.04  & 0.07   & 0.12   & 0.24   & 0.44    & 0.81    & 1.50    & 2.73     &    4.93      \\ \hline
Update all proofs time (s)          & 0.03  & 0.07  & 0.14   & 0.26   & 0.50   & 0.94    & 1.84    & 3.48    & 6.65     &    12.79      \\ \hline
\end{tabular}
\caption{Single-threaded experimental evaluation of BLTC. \textit{Update commitment time} and \textit{Update all proofs time} is the total time costs for 1024 changes to the vector. For \textit{Update all proofs time}, dividing the total time by 1024 gives an average time per update of 0.03ms to 12.50 ms.
}
\label{tab:3}
\end{table*}

\begin{table*}[]
\setlength{\tabcolsep}{1.3mm}{
\begin{tabular}{ccccc|ccc|ccc|ccc}
\hline
\multirow{2}{*}{Time Costs}                                                           & \multirow{2}{*}{Scheme} & \multicolumn{3}{c|}{$u$=8} & \multicolumn{3}{c|}{$u$ = 16} & \multicolumn{3}{c|}{$u$ = 32} & \multicolumn{3}{c}{$u$ = 64} \\ \cline{3-14} 
  &   & $n$=3   & $n$=4   & $n$=5  & $n$=3    & $n$=4    & $n$=5   & $n$=3    & $n$=4    & $n$=5   & $n$=3   & $n$=4    & $n$=5   \\ \hline
\multirow{2}{*}{\begin{tabular}[c]{@{}c@{}}Aggregate \\ proofs time (s)\end{tabular}} & Merkle SNARK  & 0.32    & 0.38    & 0.44   & 0.63     & 0.76     & 0.86    & 1.27     & 1.49     & 1.73    & 2.54    & 2.98     & 3.44    \\
   & BLTC  & 0.02    &   0.02      &  0.03      &  0.03        &   0.03       &  0.04       & 0.05   &  0.06       &  0.07       &   0.08      &    0.10      &   0.14      \\ \hline
\multirow{2}{*}{Verify time (s)}                                                       & Merkle SNARK                   & 0.003    & 0.003    & 0.003   & 0.003     & 0.003     & 0.003    & 0.003     & 0.003     & 0.003    & 0.003    & 0.003     & 0.003    \\
  & BLTC                    & 0.01    &  0.01       &  0.01      &   0.02       & 0.02         &  0.02       &    0.03        &   0.03     &   0.04      &   0.05      &   0.05       &   0.06      \\ \hline
\multirow{2}{*}{\begin{tabular}[c]{@{}c@{}}Commitment \\ time costs (s)\end{tabular}} & Merkle SNARK                   & 11.88   & 18.32   & 19.30  & 22.99    & 36.39    & 38.02   & 44.30    & 68.59    & 71.67   & 82.90   & 131.32   & 138.17  \\
  & BLTC    & 0.65    &  0.70    &  0.82      &   1.23       &   1.24       &  1.50       &    2.27   &  2.43     &  2.86       &   4.10      &   4.59       &    5.51     \\ \hline
\end{tabular}
}
\caption{Merkle SNARK scheme versus BLTC scheme. For BLTC, the aggregate or update efficiency is not affected by the watermark.}
\label{tab:2}
\end{table*}

\subsection{Performance Analysis}
we compared our proposed BLTC scheme with other existing verifiable computing schemes (\textit{Merkle} \cite{merkle1987digital}, \textit{Verkle} \cite{kuszmaul2019verkle}, \textit{Merkle SNARK} \cite{grassi2021poseidon}, \textit{Pointproof} \cite{gorbunov2020pointproofs}) in terms of performance. 
We elaborated the theoretical analysis of commitment performance from three dimensions and summarize it in Table \ref{tab:1}. 

\textit{Merkle}, \textit{Verkle} and \textit{Merkle SNARK} are verifiable computing schemes with tree structure. The proof size for the above scheme depends on the length of the submitted weight vector.
Our proof involves algebraic hashes of log$l$, so the individual proof size is the same as the Merkle tree scheme hashed with SHA-256.
Despite the same individual proof size, neither \textit{Merkle}-based commitment nor \textit{Verkle}-based commitment scheme supports the aggregation of proofs.
Although \textit{Merkle SNARK} can realize aggregation via SNARK-friendly hash functions, it is not efficient with $O(k$log$n$log$(k$log$n))$ time costs of proof aggregation. 
Therefore, the above scheme is not practical in the context of DML ownership protection. 

Compared with \textit{Pointproof}, our scheme utilizes IPA to have $O($log$n)$ update time of commitment aggregation. Since model training involves multiple weight update processes, our scheme can effectively reduce the cost of worker commitment updates. Moreover, to guarantee non-stealability, the BLTC scheme introduces an identity watermark in the proof generation process, which prevents attackers from stealing other workers’ commitments by removing signatures. 
BLTC is simultaneously aggregatable, maintainable, and non-stealable, which can effectively protect workers' ownership of the final model.


\section{Experiments} \label{sec:ex}

\textbf{Experiment Configuration.} 
To simulate the process of a model weight training, we set a series of the number of aggregated individual proof to $u=[4,$ $8,$ $16,$ $32,$ $64,$ $128,$ $256,$ $512,$ $1024]$, while setting BLT and Merkle heights to $n=[3,$ $4,$ $5,$ $30]$.
For the generation of public parameters during initialization, our BLTC scheme adopts the paring-based elliptic curve BLS12-381 to support 128-bit security. Serialized $\mathbb{G}_{1}$, $\mathbb{G}_{2}$ and $\mathbb{G}_{T}$ elements take 48, 96, and 576 bytes, respectively.
Our BLTC scheme is implemented in Golang language bindings of the \verb|mcl| cryptography library. Moreover, we realize an IPA-based proof aggregation refers to \textit{Hyperproofs} \cite{srinivasan2022hyperproofs}.
The BLTC scheme running environment of our experiment is on the Intel(R) Xeon(R) Gold 6150 CPU@2.70GHz with 18 cores and 256GB memory.
We run 10 times for each experiment and record their average.

We notice that in the case of $n=30$, the public parameter size of the \textit{Merkle SNARK} is about 50G, and it takes more than 30 hours to generate the parameters. Due to the overwhelming time costs of the SNARK-based commitment scheme, we make a trade-off on the value of $l$ and $n$. For our BLTC scheme, we fix the height of BLT to $n$ $=$ $30$ and evaluate the time costs of BLTC under $u=[4,8,16,32,64,128,256,512,1024]$. For \textit{Merkle SNARK}, we change the Merkle tree height from $n=3$ to $n=5$ in the case of $u=[8,16,32,64]$, respectively.  






\textbf{Overall Results.}
Our BLTC scheme runs single-threaded.
The efficiency of our solution can be measured by the time cost of multiple algorithms. Table \ref{tab:3} shows the time costs of our scheme with different values of $l$. We evaluate the scheme from the following three aspects:
\begin{enumerate}
    \item the time complexity of $\mathsf{BLTC.Aggr(\cdot)}$, which denotes the time cost of aggregating proofs;
    \item the time complexity of $\mathsf{BLTC.Verify(\cdot)}$, which includes the time costs of verifying individual proofs and aggregated proofs;
    \item the time complexity of $\mathsf{BLTC.ComUpdate(\cdot)}$ and $\mathsf{BLTC.}$ $\mathsf{Proof}$ $\mathsf{AllUpdate(\cdot)}$, which denote the time costs of updating commitments and updating all individual proofs respectively.
\end{enumerate}

The aggregation, verification, and update time costs of BLTC will increase as the size of the vector increases, but the corresponding time overhead is within an acceptable range. Updates to vector commitments are often involved in practical distributed training, and our scheme achieves compact vector sizes and efficient updates.

\textbf{Compared with Merkel SNARK.} 
We compared BLTC to \textit{Merkle SNARK} and summarize the experiment results in Table \ref{tab:2}. We compare a Rust implementation of \textit{Bellman-Bignat} \cite{ozdemir2020scaling}, a Merkle tree scheme that allows proof aggregation via state-of-the-art SNARKs. Moreover, to reduce the time cost of vector commitments, \textit{Merkle SNARK} is constructed from typical \textit{Pedersen} hash functions with 2753 constraints.

We evaluate BLTC and \textit{Merkle SNARK} in terms of \textit{Aggregation proofs time}, \textit{Verify time} (for an aggregated proof), and \textit{Commitment time costs} (total time for the algorithm). The \textit{Aggregation proofs time} in SNARK is primarily determined by the execution of multiple multi-exponentiations and Fast Fourier Transforms (FFTs), with the size of FFTs being linearly proportional to QAP constraints. Therefore, SNARK cannot realize effective aggregation. 
Our BLTC scheme demonstrates significantly higher efficiency in aggregation compared to \textit{Merkle SNARK}. 
In the case of the same depth $n$ of the tree, BLTC with IPA-based prover only requires computing $O(un)$ pairings and $O(un)$-sized $\mathbb{G}_{1}$, $\mathbb{G}_{2}$ and $\mathbb{G}_{T}$ exponentiations to aggregate $u$ proofs. On average, the aggregation efficiency of BLTC has increased by about 90\%.

In addition, the verification of \textit{Merkle SNARK} requires 3 pairings and $O(2u+1)$-sized $\mathbb{G}_{1}$ multi-exponentiations, while the $\mathcal{V}$ input the commitment and the size of $u$ leaves to be verified. In the case of aggregating $u=64$ proofs under $n=5$, it takes 0.003s to verify a SNARK proof and 0.06s to verify an aggregated proof in BLTC. 
Despite the increased verification cost of BLTC, 
the reduction in \textit{Commitment time costs} of BLTC reflects a substantial enhancement in overall performance,
considering the total time required for commitment and verification.
Therefore, \textit{Merkle SNARK} is ineffective in supporting DML training compared to BLTC, primarily due to the significant time cost involved in aggregating proofs and commitments.


\section{Related Work} \label{sec:rw}
\textbf{Distributed Machine Learning Security.}One of exploration directions for protecting model ownership is model watermarking. 
To protect the model copyright, the model owner embeds a watermark into the model which is unknown to the attackers. 
For example, Adi et al. \cite{adi2018turning} propose a backdoor and commitment scheme-based public watermarking algorithm, which cryptographically models deep neural networks without losing the credibility of the proof. 
SSLGuard \cite{cong2022sslguard} is a watermarking scheme specifically developed for pre-trained encoders in self-supervised learning. It introduces watermarks that contain both validation dataset and secret vector information, thereby preserving the integrity of the clean encoder while embedding the desired watermark. 
IPGuard \cite{cao2021ipguard} utilizes data samples in the proximity of the decision boundary of the original model as trigger data.
Wen et al. \cite{wen2023function} couples enhanced watermarking with the DNN model, utilizing watermark triggers generated from original training samples to combat retraining forgetting.

However, model watermarking technology usually has a trade-off between practicability and robustness, and requires corresponding modification of the training algorithm or model structure. 
In addition, model stealing attacks may remove watermarks or invalidate watermarks to infringe model copyrights \cite{jia2021entangled, wang2019neural,sun2023denet}. To protect model ownership from model stealing attacks, Guan et al. \cite{guan2022you} introduce sample correlation into model ownership protection to identify model stealing attacks by correlation difference as a robustness indicator. 
Li et al. \cite{li2022defending} identify victim models via exogenous feature embedding and meta-classifier training. 
Combining improvements in proxy networks, Mazeika et al. \cite{mazeika2022steer} propose a gradient redirection defense that alters the trajectory of extraction attacks.
Our model integrates a cryptographic vector commitment scheme and identity keys into the model training process, effectively ensuring computational integrity while incorporating the advantages of model watermarking and encryption mechanisms. 


\textbf{Verifiable Computing in Cryptography.}
Verifiable Computing aims to outsource computing tasks to third-party computing power providers.
A binary relational SNARK system involves a \textbf{prover} \verb|Prove|$(S,J)$ and a \textbf{verifier} \verb|Verify|$(S,\pi)$, where $S$ denotes a valid public input, $J$ denotes a private witness, and $\pi$ denotes a succinct generated by \textbf{prover}.
The \textbf{verifier} accepts valid proofs with absolute probability and consistently rejects any invalid proofs. The zero-knowledge property of zkSNARK ensures privacy of the witness $J$.
Verifiable Computing aims to outsource computing tasks to third-party computing power providers. 
The (untrusted) third-party computing power providers need to complete the computing task and also need to provide proof of the correctness of the computing result \cite{gowal2019scalable, xie2019libra}. 
For example, 
Niu et al. \cite{9247447} propose \textit{MVP}, a verifiable and privacy-preserving machine learning method that maintains function privacy through oblivious evaluation and batch result verification.
\textit{Drynx} \cite{9019831} utilizes homomorphic encryption, zero-knowledge correctness proofs, and differential privacy to ensure model audibility in a strong adversarial model.
In addition, verifiable computing can also be used for verifiable federated learning.
Guo et al. \cite{9285303} propose \textit{VeriFL}, a verifiable aggregation protocol with dimension-independent communication cost and bounded computational overhead.

There are some recent research works exploring the feasibility and application value of verifiable computing in practical applications. 
For example, in terms of privacy protection, researchers have proposed a computing model based on matrix commitment, which allows computer cloud servers to perform user computing tasks while protecting the privacy and security of user data. In terms of currency anonymity, the researchers also proposed a matrix commitment-based currency anonymity protocol, which can effectively protect the privacy of transaction records.
While verifiable computing can effectively ensure the computational integrity of workers, it faces challenges such as limited efficiency and overwhelming overhead during the commitment process. Some SNARK-based commitment scheme \cite{ozdemir2020scaling, lee2020replicated, tomescu2020towards} with optimistic security and reliability cannot realize effective maintenance and aggregation. 
Considering the above challenges, our model introduces tree structure and inner product arguments, which effectively constrain the aggregation cost, update cost and verification cost of the commitment process.

\section{Conclusion} \label{sec:conclu}

In this paper, we introduce a novel BLTC-DOP model to address ownership disputes and ensure computational integrity in DML. The proposed vector commitment scheme offers limited overhead and concise proofs, making it suitable for frequent parameter updates in training. By employing a maintainable tree structure, the commitment scheme reduces the cost of updating proofs. 
Another major advantage of BLTC-DOP is non-stealability via watermarked proofs. The use of worker identity keys in watermarking the proofs enhances security by preventing forgery or duplication of commitments. 
Performance analysis and comparison with SNARK-based hash commitments demonstrate the effectiveness of the binary linear tree commitment scheme in maintaining computational integrity in distributed machine learning environments.

\bibliographystyle{ACM-Reference-Format}
\bibliography{sample-base}


\end{document}